\documentclass{article}
\usepackage{fullpage}
\usepackage[round, sort]{natbib}
\usepackage{fancyhdr}
\usepackage[format=plain,
            labelfont={it,it},
            textfont=it]{caption}
\usepackage[stable, bottom, flushmargin]{footmisc}
\usepackage{graphicx}
\usepackage[pagebackref, colorlinks=true, linkcolor=blue, citecolor=cyan]{hyperref}

\usepackage{microtype}
\usepackage{graphicx}
\usepackage{booktabs} 

\usepackage[utf8]{inputenc}
\usepackage[T1]{fontenc}

\usepackage{amsmath}
\usepackage{amssymb}
\usepackage{mathtools}
\usepackage{amsthm}

\usepackage{blkarray}
\usepackage{bbm}
\usepackage{mathtools}
\usepackage{complexity}
\usepackage{makecell}
\usepackage{bm}
\usepackage{subfig}
\usepackage{titling}
\usepackage{graphicx}
\usepackage{graphics}
\usepackage{listings}
\usepackage{booktabs}
\usepackage{multirow}
\usepackage{xcolor}
\usepackage{comment}
\usepackage{amsfonts}

\usepackage{xspace}
\usepackage{thmtools} 
\usepackage{thm-restate}

\usepackage{balance}
\usepackage[capitalize,noabbrev]{cleveref}

\declaretheorem[name=Proposition]{proposition}

\declaretheorem[name=Axiom, sharenumber=proposition]{axiom}
\declaretheorem[name=Definition, sharenumber=proposition]{definition}

\declaretheorem[name=Remark, sharenumber=proposition]{remark}

\newcommand{\X}{\mathcal{X}}
\newcommand{\Y}{\mathcal{Y}}

\renewcommand{\D}{\mathcal{D}}

\renewcommand{\M}{M}

\renewcommand{\R}{\mathbbm{R}}

\newcommand{\algdet}{\mathcal{A}_\text{det}}
\newcommand{\algrand}{\mathcal{A}_\text{rand}}
\newcommand{\algthom}{\mathcal{A}_\text{thom}}
\newcommand{\algmix}{\mathcal{A}_\text{mix}}
\newcommand{\pref}{\mathbf{P}}

\newcommand{\fm}{\ell} 
\newcommand{\FM}{L} 
\newcommand{\tfm}{q} 
\newcommand{\TFM}{Q} 
\newcommand{\tam}{\widehat{q}} 
\newcommand{\TAM}{\widehat{Q}} 
\newcommand{\usf}{\tilde{\ell}} 
\newcommand{\nsf}{\widehat{\ell}} 
\newcommand{\mpm}{p} 
\newcommand{\MPM}{P} 

\newcommand{\lpfair}{\ensuremath{\textrm{OPT-LP}_\text{Fair}}\xspace}
\newcommand{\lpapprox}{\ensuremath{\widehat{\textrm{OPT-LP}}_\text{Fair}}\xspace}

\newcommand{\mv}{\ensuremath{V}\xspace}
\newcommand{\outcome}{\ensuremath{z}\xspace}
\newcommand{\outcomes}{\ensuremath{\mathcal{Z}}\xspace}
\newcommand{\Stat}[1]{\ensuremath{s_{#1}}\xspace}
\newcommand{\stat}[2]{\ensuremath{\Stat{#1}(#2)}\xspace}

\newcommand{\Mdet}{\M_\text{det}}

\providecommand{\SET}[1]{\ensuremath{\left\{ #1 \right\}}\xspace}

\providecommand{\Kth}[1]{\ensuremath{{#1}^{\rm th}}}

\providecommand{\PROB}{\ensuremath{\mathbbm{P}}\xspace}
\providecommand{\Prob}[2][]{\ensuremath{%
\ifthenelse{\equal{#1}{}}{\PROB[#2]}{\PROB_{#1}[#2]}}\xspace}

\providecommand{\Expect}[2][]{\ensuremath{%
\ifthenelse{\equal{#1}{}}{\mathbb{E}}{\mathbb{E}_{#1}}%
\left[#2\right]}\xspace}

\newcommand{\LPlabel}{}
\newenvironment{LP}[3][]{%
\renewcommand{\LPlabel}{#1}
\ifthenelse{\equal{\LPlabel}{}}{%
\[ \begin{array}{ll}
\mbox{#2} & \;\;#3 \\
\mbox{\normalfont s.t.} & \begin{array}[t]{ll}%
}{%
\begin{equation} \begin{array}{ll}
\mbox{#2} & \;\;#3 \\
\mbox{\normalfont s.t.} & \begin{array}[t]{ll}%
}}{%
\ifthenelse{\equal{\LPlabel}{}}{%
\end{array} \end{array} \]}{%
\end{array} \end{array} \label{\LPlabel} \end{equation}}%
}

\theoremstyle{plain}
\graphicspath{{arxiv_draft_2_content/}}

\newcommand\blfootnote[1]{%
  \begingroup
  \renewcommand\thefootnote{}\footnote{#1}%
  \addtocounter{footnote}{-1}%
  \endgroup
}

\title{Fairness in Matching under Uncertainty}

\author{
   Siddartha Devic\footnotemark[1]
   \and
   David Kempe\footnotemark[1]
   \footnotemark[3]
   \and
   Vatsal Sharan\footnotemark[1]
   \footnotemark[3]
   \and
   Aleksandra Korolova\footnotemark[2]
   \footnotemark[3]
}

\begin{document}
\maketitle{}

\blfootnote{\hspace{-4pt} {*} \hspace{-4pt} Department of Computer Science, University of Southern California, Los Angeles, CA 90089. Emails: \texttt{devic@usc.edu}, \texttt{david.m.kempe@gmail.com}, \texttt{vsharan@usc.edu}.}

\blfootnote{{\dag} \hspace{-1.3pt} Department of Computer Science and School of Public and International Affairs, Princeton University, Princeton, NJ 08544. Email: \texttt{korolova@princeton.edu}.} 

\blfootnote{{\ddag} \hspace{-1pt} Equal contribution; the order of these authors was randomized with \url{https://www.random.org/}.}

\begin{abstract}
The prevalence and importance of algorithmic two-sided marketplaces has drawn attention to the issue of fairness in such settings. 
Algorithmic decisions are used in assigning students to schools, users to advertisers, and applicants to job interviews.
These decisions should heed the preferences of individuals, and simultaneously be fair with respect to their merits (synonymous with fit, future performance, or need).
Merits conditioned on observable features are always \emph{uncertain}, a fact that is exacerbated by the widespread use of machine learning algorithms to infer merit from the observables.
As our key contribution, we carefully
axiomatize a notion of individual fairness in the two-sided marketplace setting which respects the uncertainty in the merits; indeed, it simultaneously recognizes uncertainty as the primary potential cause of unfairness and an approach to address it. 
We design a linear programming framework to find fair utility-maximizing distributions over allocations,
and we show that the linear program is robust to perturbations in the estimated parameters of the uncertain merit distributions, a key property in combining the approach with machine learning techniques.
\end{abstract}

\section{Introduction}
\label{sec:introduction}
Systems based on algorithms and machine learning are increasingly used to guide or outright make decisions which strongly impact human lives; thus it is  imperative to take \emph{fairness} into account when designing such systems.
Notions of fairness in computer science can be classified into those that try to capture fairness towards a \emph{group} 
\citep{hardt2016equality, kleinberg2016inherent, kearns2018preventing,
hebert2018multicalibration}
vs.~those that try to be fair to each \emph{individual} \cite{dwork2012fairness, kim2018fairness, kim2020preferenceInformedFairness}.
In our work, we focus on the latter notion.
The most widely studied notion of individual fairness is due to the seminal work of \citet{dwork2012fairness}: it assumes that a metric space on observable features of individuals captures similarity, and requires that outcomes of  a resource allocation mechanism satisfy a certain Lipschitz continuity condition with respect to the given metric. 
Intuitively, this ensures that individuals who are \emph{similar} according to the metric will be treated \emph{similarly} by the mechanism. 

We consider a setting in which individuals have \emph{preferences} over the outcomes of the resource allocation mechanism, focusing on the important
setting of two-sided markets. Applications of this setting abound: matching students to schools, job fair participants to interviews, doctors to hospitals, patients to treatments, drivers to passengers in ride hailing, or advertisers to ad slots/users in online advertising \citep{roth1986allocation, abdulkadirouglu2003school, roth2007efficient, bronfman2015assigning,mehta2013online}, to name a few. 
In all these settings, the \emph{individuals} (students, doctors) can communicate preferences over the \emph{resources} (schools, hospitals), and also have observable features which may (partially) reveal their qualifications or \emph{merit} for the different resources. Merit is broadly construed and can be defined as a myriad of things: the true capabilities of a job candidate, the potential relevance of a doctor's interests when matching with a residency program, a patient's need for treatment, 
or the fit of a student applying for public high schools. 
The goal of our work is to develop a framework for reasoning about fairness in such settings, as well as an algorithm which assigns individuals to resources in a way that is fair with respect to their merit, but also takes individual preferences into account. 
In addition, the algorithm should attempt to maximize the overall utility of the system (according to problem-dependent notions of overall utility, e.g., social welfare/revenue), while respecting fairness constraints.

Reconciling individual preferences with the traditional notion of individual fairness (IF) has proven to be difficult. 
The most relevant approach is that of \citet{kim2020preferenceInformedFairness}, who propose \emph{preference-informed individual fairness} (PIIF), which requires that allocations be individually fair but also allows for deviations aligned with preferences of the users.
However, there is a key difference between \emph{allowing} for deviations in line with user preferences and \emph{requiring} a stronger notion of fairness in the presence of preferences.
As an example, consider the case of matching two applicants $A$ and $B$ to jobs $X$ and $Y$. Suppose
$A$ and $B$ have similar qualifications, but $A$ strictly prefers $X$ over $Y$, and $B$ strictly prefers $Y$ over $X$.
In this setting, it is clear that we should assign $A$ to $X$ and $B$ to $Y$: both $A$ and $B$ are equally qualified, and they can both receive their top choice.
Furthermore, doing so does not even come at a cost to $X$ and $Y$ since $A$ and $B$ are essentially indistinguishable in terms of merit.
Nonetheless, PIIF allows the following (randomized) allocation: assign $A$ to either $X$ and $Y$ by flipping a coin, and assign $B$ to the remaining job.
This allocation is \emph{individually fair} (since both $A$ and $B$ have identical distributions over outcomes); by extension, it is also PIIF.

We believe that fundamentally, the central axiom of individual fairness (``individuals with \emph{similar} features must be treated \emph{similarly}'') is incompatible with individual preferences (``if individuals have different preferences, each can get their top choice'').
This tension disappears when instead of viewing qualifications as necessitating similar treatment, we view them as entitlements to desirable outcomes. These entitlements should depend not only on the individuals' absolute qualifications and preferences in a vacuum, but their qualifications and preferences in the \emph{context} of all other individuals and resources. We regard this as the \emph{contextual entitlement} of an individual to obtain their desired outcome.

To define the appropriate notion of contextual entitlement and ascertain stronger fairness guarantees than applying IF or PIIF directly, we build on the recent work of \citet{singh2021fairness}, who question the Lipschitz continuity requirement on the algorithm's allocation imposed by individual fairness. They argue that Lipschitz continuity should not be treated as a first-order desideratum (optimized directly over), but rather as a derived consequence of the \emph{uncertainty} involved in estimating an individual's true merit based on their observed features.
Specifically, their central thesis is that merit is never perfectly captured by observable features; therefore, the role of observable features is to induce a posterior joint merit distribution over individuals. 
We note that in numerous contemporary matching markets, ML algorithms 
are being used to infer the merits or relevance of the individuals to the resources. For example, ML algorithms are pervasively used in internet advertising to determine the relevance of an advertiser to an ad slot/user \citep{mcmahan2013ad, linkedInRelevanceScores, fb2023vrspaper} and are increasingly common in recruiting platforms to determine the suitability of a candidate for a job \citep{schumann2020we}. The outputs of these algorithms are inherently uncertain; in fact, many ML algorithms output probability estimates, and many others can be modified to do so 
\citep{smith2013uncertainty}.

Treating uncertainty as the keystone of fairness, \citet{singh2021fairness} argue that the intuitive reason why similar individuals should be treated similarly (the requirement of individual fairness) is that \emph{similar observable features} give rise to \emph{similar posterior distributions over merits}.
Using this intuition, they derive a notion of fairness which states that when each individual has higher merit with roughly equal probability, each should be treated better with roughly equal probability. More generally, the work of \citet{singh2021fairness} considers the setting of choosing a \emph{ranking} of individuals, e.g., ranking search results on an e-commerce site. Their approach can be viewed as defining a notion of entitlement of a seller to be displayed prominently in that setting.

\paragraph{Contributions.} 
In this work, we build on the work of  \citet{singh2021fairness} to derive suitable notions of contextual entitlement of an individual to get their desired outcome in bipartite matching with individual preferences. 
Our primary contribution is the definition (in \cref{sec:preliminaries}) of the framework of uncertainty, and the axiomatization of a suitable notion of individual fairness \emph{with preferences} due to and under uncertainty.
Prior work appeared to struggle to reconcile individual preferences with fairness even in this basic setting, and we demonstrate that a focus on uncertainty and merits resolves this issue while paving a path forward for approaching fairness in more general allocation problems widely studied in the context of ML. 

\begin{figure*}[ht]
\begin{centering}
  \includegraphics[scale=0.25]{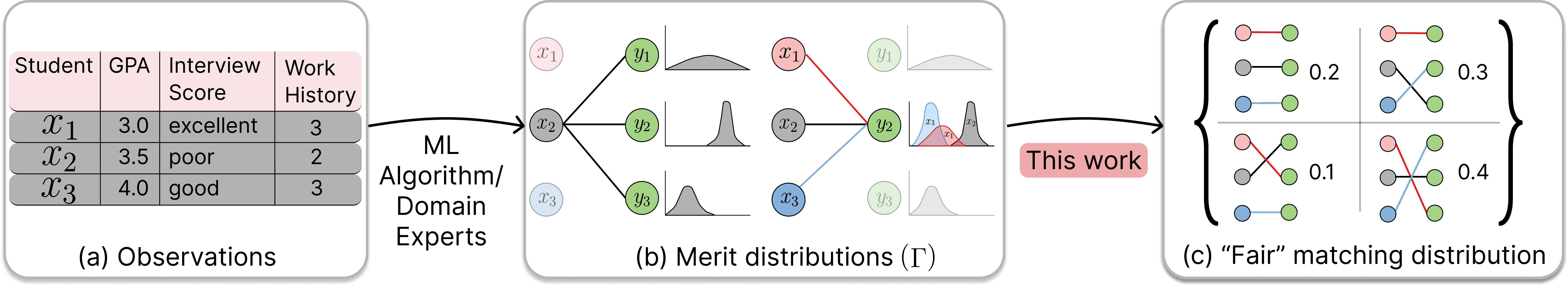}
  \caption{
  A visual overview of our setup, using matching \emph{students} and \emph{jobs} as an example.
  Each job has an ML algorithm or domain expert with access to observed features (a) for each student, and from those constructs a distribution $\Gamma$ 
  (b) which is an estimate of the merit or \emph{fit} for the job/student pair.
  Importantly, these estimates could be based on different factors for different jobs (see heterogeneity of merit distribution for student $x_2$).
  Student preferences (not pictured) and $\Gamma$ are given to the algorithm as input. 
  The algorithm seeks to output a distribution over matchings which is ``fair'' (c) w.r.t.~what each student is contextually entitled to, while simultaneously considering the preferences of \emph{all} students and the overall utility of the solution.
  }
  \label{fig:main_fig}
\end{centering}
\end{figure*}

Our technical contributions (detailed in \cref{sec:results}) are to explore the resulting fairness/utility trade-offs in the setting of two-sided matchings, rather than merely rankings (which were examined in \citet{singh2021fairness}).
We present an algorithm whose output is a probabilistic matching of the individuals to resources (see \cref{fig:main_fig}).
When the principal's utility can be expressed as a sum of the utilities of the matches between individuals and resources, we show that the principal's optimization problem can be cast as a linear program (LP).
Our fairness framework enables efficient algorithms for satisfying fairness while still maximizing the overall utility of the system, and also allows for \emph{tradeoffs} between utility and fairness. 
Our primary technical contribution is to show that small errors in the principal's estimates of the entitlements only lead to a small loss in utility and fairness; as a result, even when entitlements are estimated using samples, the number of samples required to ensure a utility/fairness loss of at most a $(1-\epsilon)$ factor grows only polynomially in $1/\epsilon$.
Our contributions offer a new lens for considering tradeoffs between a principal's desiderata and fairness, and thus open avenues for future work (\cref{sec:future-work}).

\section{Preliminaries}
\label{sec:preliminaries}
\subsection{The Setting and Fairness Principles}
There are $n$ \emph{individuals} (such as students, doctors, patients) $\X$ and $n$ \emph{resources} (such as jobs, residency positions, hospital beds) $\Y$.
Each individual is to be matched to exactly one resource; we discuss generalizations in \cref{sec:extensions}.
Each individual $x$ has a (total) preference \emph{ranking} over resources, captured by the bijection $r_x : \Y \to [n]$.
Thus, $r_x^{-1}(k)$ is the \Kth{k} choice of individual $x$;
in particular,  $r_x^{-1}(1)$ is the top choice of $x$.
These rankings are communicated to the algorithm by the individuals, and thus treated as deterministic.

In addition, individuals have observed features, such as GPA, performance on standardized tests, performance in job interviews, medical tests, documentation of needs, etc. 
These observed features reveal (partial) information about the merits of the individuals for the resources.
The merit of individual $x \in \X$ for resource $y \in \Y$ is denoted by $v_{x,y}$, and $V = (v_{x,y})_{x \in \X, y \in \Y}$ is the matrix of all merits.
We assume that the merits are unknown, but are drawn from a known distribution $\Gamma$; this distribution is estimated by experts or an ML algorithm based on the individuals' observable attributes.
We assume that ties in merit have probability 0, i.e., for any $y$ and $x \neq x'$, the event $v_{x,y} = v_{x',y}$ has probability 0.\footnote{This condition states that no two individuals ever have the \emph{exact same} merit for any resource. This is ensured with probability 1 for example if each entry has the tiniest amount of independent noise drawn from a continuous distribution. Note that we are assuming this only for the individuals' \emph{merits}, not their observable \emph{features}. For the latter, we may well be in a situation where only students' GPAs are known, and several have identical GPAs.} 
We make no assumptions about $\Gamma$; e.g. we allow arbitrary correlations between the $v_{x,y}$ values.
For the entirety of this paper, the observable attributes are fixed, so we will
just focus on the known distribution $\Gamma$.
Our model follows \citet{singh2021fairness}; it ascribes no  semantics or meaning to the observed features, aside from inducing a distribution.
In particular, we do not assume that the observed features are numeric (e.g., they may be verbal summaries of a job interview or a written personal statement), or that a metric can be defined over them.

Although there is uncertainty in the merits of the individuals, a mechanism designer, or the \emph{principal}, should still strive to ensure that individuals receive ``good'' or \emph{fair} outcomes.
While our ultimate goal is to obtain a suitable definition of fairness under uncertainty, we begin with an axiom for deterministic fairness in matching, i.e., under \emph{absolute certainty}: 

\begin{axiom}[Fairness of a Matching with Certain Merits]
\label{axiom:fairness_certainty}
Assume that all merits $v_{x,y}$ are perfectly known, and that there are no ties, i.e., $v_{x,y} \neq v_{x',y}$ for all resources $y$ and all $x \neq x'$.
Let $M: \X \to \Y$ be a matching (bijection) from individuals to resources.
We say that \emph{$M$ is fair towards individual $x$} if for all resources $y$, $v_{M^{-1}(y),y} < v_{x,y}$ implies that $r_x(y) > r_x(M(x))$.
In words, if resource $y$ goes to an individual with less merit than $x$, then $x$ obtains a resource she prefers over $y$. (Otherwise, $y$ would go to a less qualified individual, while a more qualified individual would be worse off.)
$M$ is \emph{fair} if it is fair towards all individuals $x$.
\end{axiom}

\cref{axiom:fairness_certainty} expresses an extreme notion of \emph{meritocracy}.
It articulates that if all merits were indeed  known, then the resources should always go to more qualified individuals if they desire them.
Such an extreme meritocratic approach may be startling, and appear unfair.
However, notice that the axiom is fully predicated on the assumption that merits are perfectly known.

In reality, as articulated above, merits will be only imperfectly predicted by observable features, and the resulting uncertainty about which individual has higher merit should be seen as the true force towards more equal treatment.
Doing so avoids unprincipled decisions about how similarly individuals should be treated based on ``small'' or ``medium'' differences in merit, and in effect implies that merits are only used ordinally.

As discussed in depth in \cref{sec:discussion}, any allocation rule that aims to treat similar individuals ``similarly'' must randomize the outcomes. The key question is how to derive the probabilities in a principled way from a normative axiom. We base our approach on the axiom that individuals should not be punished for the mechanism's uncertainty in the merits. This is captured formally by the following \emph{lifting axiom}.

\begin{axiom}[Lifting Axiom]
\label{axiom:lifting}
Consider a setting of decision making under uncertainty (not necessarily matching), in which all relevant merits $\mv$ are drawn jointly from a known distribution $\Gamma$.
Let $\algdet: \mv \mapsto \outcome \in \outcomes$ be a fair deterministic algorithm mapping the merits $\mv$ to a corresponding outcome $\outcome$, satisfying a suitable notion of deterministic fairness.
Let $\Stat{1}, \Stat{2}, \ldots, \Stat{k}: \outcomes \to \R^{\geq 0}$ be statistics on the outcomes which are deemed relevant for fairness.
Then, a randomized algorithm (i.e., a map which takes as input a distribution over merits $\Gamma$ and outputs a distribution over the outcome space $\outcomes$) 
is \emph{$\phi$-fair} (for some fairness parameter $\phi \in [0, 1]$) if it satisfies
$\Expect[\outcome \sim \algrand]{\stat{j}{\outcome}} \geq \phi \cdot \Expect[\mv \sim \Gamma]{\stat{j}{\algdet(\mv)}}$ for all statistics $j$.
\end{axiom}

\Cref{axiom:lifting} requires some ``unpacking'' and justification.
It considers as a baseline a hypothetical world in which a fair deterministic algorithm is run after the random merits have been drawn \emph{and fully revealed}.
Doing so results in a distribution over outcomes $\outcome$, where the randomness is the result of random merits.
The outcome statistics $\Stat{j}$ are the quantities which the mechanism cares about; in the case of matching, these will be each individual $x$'s probability of obtaining one of her top $k$ outcomes, for each $k \in [n]$.
Our lifting axiom, with the fairness relaxation parameter $\phi$ set to 1,
then states that just because the randomized mechanism does not have access to the true merits, it should not give an individual a worse outcome by any statistic than if it \emph{did} have access to the true merits (and ran the baseline deterministic algorithm $\algdet$).
Or, stated differently, uncertainty about merits should not be a reason to discriminate against any individual with respect to any statistic of interest.

\begin{remark}
The key difference between $\algdet$ and $\algrand$ is that the input to $\algdet$ is a matrix of deterministic merits $V$ (for example, the fully known merits of job applicants for different companies), and the output of $\algdet$ is a deterministic outcome $z$ (a matching between jobs and applicants).
In contrast, the input to $\algrand$ is a distribution over merits for each applicant for each job $\Gamma$ (which models inherent uncertainty in estimating the value of applicants), and the output of $\algrand$ may be randomized over outcomes (a distribution over matchings of applicants and jobs).
\end{remark}

The choice of statistics $\Stat{j}$ captures the desired notion of fairness. Our mathematical formulation, treating the expectations as \emph{lower bounds}, implicitly captures that larger values are better or more fair. Apart from this, we place no restriction on the $\Stat{j}$. In principle, $\Stat{j}$ could even completely omit the outcomes of some of the individuals, though the most natural approach in a symmetric setting is to apply the same statistics to the outcomes of all individuals.
Considering our approach for general environments, the statistics should be chosen by domain experts, and auditable as quantitative expressions of normative criteria of desired fairness.

As in \citet{singh2021fairness}, we consider a relaxed notion of fairness, which quantifies the degree to which a principal may be allowed to deviate from full fairness.
Such a relaxed notion is useful to study the tradeoff between fairness towards the individuals and maximizing some other notion of utility for a principal (such as the principal's revenue or some global societal goal). 
The parameter $\phi$ captures the extent to which fairness is desired, with $\phi=1$ being full fairness, and $\phi=0$ leaving the principal unconstrained.

We remark that the fairness notion/axioms of \citet{singh2021fairness} for \emph{ranking} under uncertainty are a special case of our definitions when all individuals have the same ranking $r_x = r_{x'}$ over resources, so that there is an objective best (most desired) resource, second-best, etc.
\citet{singh2021fairness} required that each individual obtain a choice in the (common) top $k$ with probability at least equal to having merit in the top $k$.
This requirement was stated somewhat ad hoc, and we believe that \cref{axiom:lifting} articulates a more general principle from which their requirement can be derived.

\subsection{The Principal's Utility}
\label{sec:principal-utility}

In addition to the criterion of fairness towards the individuals, we also consider the principal's utility.
This utility may or may not be aligned with the individuals' preferences or their merits for the resources.
We write $\mu_{x,y}$ for the utility the principal derives from giving resource $y$ to individual $x$.
Thus, the principal's utility from a matching $\M$ is $U(\M) = \sum_x \mu_{x, \M(x)}$.
If the principal is trying to maximize social welfare of the allocation to resources, i.e., the sums of merits of everyone for the resource they obtain, then a natural choice is $\mu_{x,y} = \Expect[V \sim \Gamma]{v_{x,y}}$; however, we allow generic $\mu_{x,y}$.
For most of our results, we will assume that $\mu_{x,y} \geq 0$ for all $x, y$. This is solely because multiplicative approximation (of utility) --- the most natural and widely used approach --- becomes meaningless or impossible when the objective function could be positive or negative.

A principal constrained by fairness desiderata will face a tradeoff between his utility and the fairness towards individuals. We now capture the principal's optimization problem of maximizing his utility under a given fairness requirement:
\begin{definition}[Utility maximizing $\phi$-fair matching]
  Given a desired fairness level $\phi$, the principal seeks a randomized allocation algorithm $\algrand$ maximizing $U(\algrand) := \sum_{x,y} \mu_{x,y} \cdot \Prob[\M \sim \algrand]{\M(x) = y}$, subject to the constraint that $\algrand$ is $\phi$-fair, i.e., satisfies \cref{axiom:lifting}.
\end{definition}

Note that by linearity, the principal's utility can be fully expressed in terms of the marginal probabilities with which each individual is assigned each resource.
Thus, if $\MPM = (\mpm_{x,y})_{x,y}$ is a doubly stochastic\footnote{Recall that a matrix is doubly stochastic if all entries are in $[0,1]$ and each row and column sums to 1.} matrix with $p_{x,y}$ representing the probability that individual $x$ is matched with resource $y$, we also write $U(\MPM) = \sum_{x,y} \mu_{x,y} \cdot \mpm_{x,y}$ for the principal's expected utility under the marginal probabilities $\MPM$.
The \emph{Birkhoff von Neumann} (BvN) decomposition \citep{birkhoff1946tres} gives a constructive and efficient way to decompose any doubly stochastic matrix into a convex combination of permutation matrices (i.e., matchings), so the desired distribution over matchings can indeed be efficiently obtained from $P$.

\subsection{Characterization of Fairness via Stability}
In order to apply \cref{axiom:lifting} to the case of matching under uncertainty, we require the fairness notion of \cref{axiom:fairness_certainty} for the deterministic algorithm $\algdet$.
Our first observation is that fairness is equivalent to stability when the merits are fully known.
\begin{restatable}{proposition}{fairnessStability}
\label{prop:fairness_stability_equiv}
  Let $V$ be a matrix of merits without ties.
  For each resource $y$, let $r'_y$ be a ranking of all individuals in order of decreasing merits of each individual for $y$.
  Then, a matching $M$ is fair according to \cref{axiom:fairness_certainty} iff $M$ is a stable matching with respect to the rankings $(r_x)$ and $(r'_y)$.
\end{restatable}
\begin{proof}
  We show that $M$ is unfair towards $x$ at resource $y$ if and only if $(x,y)$ is a blocking pair.
  Then, fairness towards all individuals at all resources is equivalent to $M$ being stable.
  By \cref{axiom:fairness_certainty}, $M$ is unfair towards $x$ at resource $y$ if $v_{M^{-1}(y),y} < v_{x,y}$ and $r_x(y) < r_x(M(x))$.
  The first condition is equivalent, by definition of $r'_y$, to $r'_y(x) < r'_y(M^{-1}(y))$.
  This in turn is equivalent, by definition, to $(x,y)$ forming an unstable pair under the rankings.
\end{proof}
To capture the notion of fairness under a randomized allocation algorithm,
fix an individual $x \in \X$ and $k \in \SET{1, \ldots, n}$.
For a matching $\M$, we define the fairness statistic $\stat{x,k}{\M} = \mathbbm{1}[r_x(\M(x)) \leq k]$, i.e., the $s_{x,k}$ are indicator functions of whether individual $x$ obtained one of her top $k$ choices.
These statistics have the following useful properties: (1) together, they precisely determine the matching $\M$ (because each $x$ is matched to the unique position in her ranking satisfying $\stat{x,k}{\M} > \stat{x,k-1}{\M}$); (2) each agent's utility is (weakly) monotone in each statistic.
Property (1) ensures that, in a sense, we capture all information that can be captured; and property (2) ensures that defining fairness by requiring larger values for statistics is meaningful.

Our goal is to design an algorithm $\algrand$ which takes as input $\Gamma$ and generates a $\phi$-fair distribution over matchings. 
Using our lifting \cref{axiom:lifting}, we obtain the following equivalent definition of fairness of a randomized allocation algorithm under uncertainty which we will use throughout.
To express the definition concisely,
we write 
$\fm_{x,k} := {\Prob[\mv \sim \Gamma, \Mdet=\algdet(\mv)]{r_x(\Mdet(x)) \leq k}}$
for the probability that $x$ gets her \Kth{k} or higher-ranked resource under $\Gamma$ and when the deterministic algorithm $\algdet$ chooses the stable matching $\Mdet$ that is optimal for the individuals.
\begin{definition}
    A distribution over matchings $\algrand$ is \emph{$\phi$-fair} if for all $x \in \X, k \in [n]$, we have that
    \begin{align}
      &\underset{\mathclap{M \sim \algrand}}{\mathbbm{P}}\ \ [r_x(\M(x)) \leq k] \geq \phi \cdot \fm_{x,k}.
    \label{eqn:fairness-requirement-after-axiom}
    \end{align}
\end{definition}

For the definition, we make a normative decision to choose the stable matching which is (simultaneously) optimal for all individuals (as opposed to, say, the resources according to the rankings $r'_y$, or optimal for neither).
This is achieved by letting $\algdet$ be individual-proposing Gale-Shapley,
and is the most stringent fairness requirement, as it gives all individuals the highest possible rank of any fair allocation.
One could study trade-offs within our lifted uncertainty framework (\cref{axiom:lifting}) with respect to other matchings/choices of $\algdet$, which would be more lenient.
For example, the most lenient candidate would be a \emph{resource}-proposing stable marriage algorithm; obtaining other stable matchings besides these two natural ones is typically hard \citep{gusfield1989stable}.
However, we believe that in light of the primary focus on fairness, optimizing for the individuals is natural.

\section{Discussion and Related Work}
\label{sec:discussion}
Given that our key contribution is to articulate a formal framework for reasoning about fairness in allocation problems, before proceeding to technical results, here, we discuss various modeling choices and their underlying implicit or explicit normative principles, as well as limitations imposed by our framework.
Many of the issues we discuss here have been discussed by \citet{singh2021fairness} as well.

\subsection{Observables and Merit}

An extremely important aspect of our framework is the distinction between observable features and merits of individuals.
Observable features, as discussed previously, encompass items such as GPA, verbal summaries of interviews, application materials, scores on standardized exams, etc.
In contrast, merit is a rather amorphous notion, capturing a mixture of qualification (for jobs) or need (for resources). The units in which it may be measured, or the specific notion, may be difficult to articulate. For example, for the scenario of hiring, should merit be the immediate readiness of an applicant, the performance after one year, after five years, or some combination thereof? For access to a life-saving drug or medical treatment, should merit be an estimated probability of dying without it or should it take into account other health conditions, expected remaining lifespan, etc.?

The fact that merit is conceived as an abstract notion also allows the principal to incorporate risk aversion, or --- conversely --- risk seeking behavior. For example, if a job only requires basic competency, then merit may be defined as the probability of exceeding such a minimal competency level. Conversely, if only exceptional performance is valued, then the probability of being exceptional may be a suitable notion of merit. 
By modifying the specific definition of merit, the principal can control how he prioritizes the value of ``risky'' candidates who are, for example, exceptionally qualified with low probability and average otherwise.

The amorphous nature of merit may appear to be a drawback of our approach, since any utility one may derive is contingent upon defining a suitable and useful notion of merit. 
However, we believe that it is, in fact, a feature.
Any discussion of fairness must be underpinned by an understanding of how deserving individuals are of which resources. The notion of merit serves as a clean abstraction for articulating this entitlement.
Formulating an appropriate notion of merit should fall to domain experts, and the role of computer science within this context should be to help the principal achieve fairness with respect to the proposed merit notions.

\subsection{Merit Distributions}

The discussion of merit above, and the examples we gave, should make it clear that meaningful notions of merit are rarely if ever observable. 
For example, performance on a job after one year may be somewhat predicted by observable attributes, but will also be determined to a large extent by random future events.
Thus, our second key modeling assumption is that the observable features of all individuals only give a distribution over their merits for the resources, rather than deterministic values. 

The allocations produced by mechanisms under our framework will only be as fair as allowed by the distributions.
If the distributions do not capture actual merit based on observables, a mechanism that is ``fair'' with respect to the assumed distributions will fail to be so in a meaningful sense. 
Thus, in addition to defining a suitable notion of merit, domain experts (or ML-based predictors) will also be needed to articulate what the observable features reveal about an individual's merits for all resources.

\subsection{The Principal's Utility}
Our approach can be applied to any utility function for the principal which can be expressed in terms of utilities for individual resource assignments. (See a discussion of the difficulties with a generalization beyond this setting in \cref{sec:extensions}.)
This includes settings in which the principal's utility may be well aligned with the individuals' preferences (e.g.
when the principal derives utility whenever an individual is matched to their top choice).
We remark that such settings are in a sense ``easier'', in that the lack of conflict between the principal's and individuals' utilities leaves the principal less constrained in terms of optimization.

\subsection{Randomization}
Our model treats uncertainty of merits as a first-order feature/concern, but also ``fights fire with fire'': our algorithmic approach heavily relies on randomizing the allocation of resources to individuals.
In general, explicitly randomized allocations are not as widely used in real-world settings (outside of gambling) as deterministic ones.
They do seem to be accepted more readily when applied to repeated low-stakes settings than for rare or single-shot high-stakes ones.
Two plausible explanations suggest themselves: (1) individuals may not trust the principal's claim to randomize, and suspect that the outcome may be rigged. (2) In repeated settings, the actual average allocation will usually be close to the expectation, resulting in fairness not only \emph{ex ante}, but also \emph{ex post}.
Given the relatively lower adoption of randomization in practice (especially in high-stakes settings), a natural question is whether guarantees for approaches utilizing forms of individual fairness (including ours) can be obtained without randomization.

We argue that in general, this is impossible: while randomization of outcomes may appear undesirable, it is unavoidable when any similarity-based notion of individual fairness is to be achieved. 
This can be seen in two contexts. 
First, if there are two similar individuals, both preferring the same resource, any deterministic allocation would leave one agent always with the less desirable resource; this would constitute very dissimilar treatment.
Even in the absence of resource constraints, similar issues will arise. Consider two very different individuals who should be treated differently (e.g., one should definitely be given a loan, while the other should not). Now, following a standard proof technique, consider a sequence of individuals ``interpolating'' between the two, so that any two adjacent individuals in the sequence are similar. If allocations are made deterministically, there must be at least one adjacent pair in the sequence such that one is deterministically allocated, while the other goes deterministically unallocated. This would violate similar treatment of similar individuals. 

Naturally, whether in a particular context, randomization is an acceptable approach is beyond the purview of technical work. Instead, it should be decided by the domain expert seeking to achieve fairness, after articulating in what sense fairness is desired.
The concrete contribution of our work is to articulate a more fundamental underpinning of quantitative randomization decisions when similarity-based individual fairness is indeed desired.

\subsection{Related Work}
\label{sec:related}
Our fairness framework, as previously mentioned, is built by generalizing the uncertainty framework of \citet{singh2021fairness} to matching.
\citeauthor{singh2021fairness} introduce the concept of working with the merit distribution $\Gamma$ and comparing individuals by the probability that one is more qualified than another.
They propose an LP framework for finding approximately fair and utility maximizing randomized rankings, and also show the method is practical by fielding it in a real-world conference.
We remark that the proof of Prop.~4.4 in \citet{singh2021fairness} (analogous to our \cref{thm:main_result}) 
contains a mistake that appears to be unfixable.
Thus, an important part of our contribution is not only generalizing their result to matchings,
but also obtaining a correct and tight proof even for the restricted ranking setting.

Most similar to our setting is the work of \citet{karni2022fairnessAndStability}, focusing on producing stable matchings which also satisfy the PIIF notion of \citet{kim2020preferenceInformedFairness}.
Recall that PIIF is based on the concept of IF \cite{dwork2012fairness}, but critically allows for deviations away from IF if these deviations align with an individual's preferences.
As discussed in \cref{sec:introduction}, while PIIF is a natural generalization of IF in non-resource-constrained settings (like classification), the guarantees it provides deteriorate substantially under constrained resources.
This leads to two major differences between our work and that of \citet{karni2022fairnessAndStability}.
The most important is that our method \emph{requires} fairness with respect to a stronger baseline (namely, individual-proposing Gale-Shapley), while PIIF based approaches like \citet{karni2022fairnessAndStability} \emph{allow} for these solutions plus many other, less fair ones which potentially include worse outcomes for individuals.
The second difference is that \citet{karni2022fairnessAndStability} work in the more difficult setting of simultaneously guaranteeing PIIF and an appropriate generalization of stability. 
Indeed, they must restrict discussion to PIIF with \emph{proto}-metrics (where all distances between pairs of individuals are either 0 or 1), and extending results beyond this setting runs into difficult technical challenges and impossibility results.
In contrast, we (1) do not consider stability guarantees of our final solution\footnote{While our method indeed utilizes Gale-Shapley as a subroutine, we \emph{do not} have any stability guarantees for our final matching distribution.}; and (2) believe that our framework of contextual entitlement more naturally captures desiderata surrounding preferences and qualifications than PIIF.

Very recent work of \citet{hogsgaard2023optimally} considers theoretical guarantees of mixing fair and utility maximizing mechanisms more generally (but does not consider general user preferences).
Their approach could potentially be used to give utility guarantees on the (suboptimal) Thompson sampling approach of \citet{singh2021fairness}.
In our work, we derive guarantees based on our specific problem formulation, and therefore expect better approximation results.

\subsubsection{Additional Related Work}
We also survey some related work from the literature from two-sided matching markets, fair division, statistical discrimination, and online/offline fair ranking and matching, and note differences to our proposed framework.

\paragraph{Two-sided Markets.} Nearly all of the literature deals with the setting of fairness in one- or two-sided markets with exact (as opposed to \emph{uncertain}) preferences/utilities.
\citet{do2021two} use Lorenz efficiency to create Pareto efficient rankings, producing rankings/matchings which increase the utility of worse-off individuals.
There is also work on group fairness in bipartite matching.
\citet{panda2022bipartite} consider bipartite matching with group fairness constraints and enforce minimum/maximum selection rates per group (quotas), but do not consider preferences of either side. 
They do have a notion of individual fairness (probabilistic individual fairness), but it is still defined in terms of a distribution over \emph{group} fair deterministic matchings.
\citet{fleiner2016matroid} work with similar desiderata but in a setting with preferences on both sides, proving that lower quotas and stability are compatible. 
Similarly, using preferences, \citet{huang2016fair} define a fair bipartite matching to be one which, subject to having maximum cardinality, minimizes the number of rank-$n$ neighbors, then minimizes the number of rank-$(n-1)$ neighbors, etc. They also 
provide an algorithm for finding such a (deterministic) matching.
In contrast to these works, our model has uncertainty, and we provide individual-level statistical guarantees relative to the expectation of those statistics.

In two-sided markets more generally, \citet{su2022optimizing} study an \emph{apply-accept} interaction protocol, in which individual users are given additional autonomy over recommended content.
They empirically show that individual utilities are distributed in a more egalitarian manner if, when optimizing over preferences for individual users, the recommendation algorithm indeed considers what other users are being recommended as well (similar to our notion of contextual entitlement).
In contrast, we treat fairness as a first-order desideratum and \textit{explicitly} constrain our optimization problem to respect it.

\paragraph{Fair Division and Stable Matchings.} The fair division and mechanism design literature
\citep{Amanatidis2022FairDivision, cole2013mechanism} has other approaches for dealing with resource allocation; however, they typically do not consider the existence of a classifier or underlying merit distribution $\Gamma$.
In particular, we also take inspiration from works that consider an unknown amount of good to be divided \citep{xue2018fair, long2021equal} since they too deal with uncertainty.
There is also work approaching fairness in matching from the perspective of the fair division literature.
\citet{freeman2021twosided} consider the notion of ``Double Envy-Freeness Up To $c$ Matches'', which ensures that individuals are satisfied up to some number of resources being removed from the picture.
This is motivated by envy-freeness up to $c$ goods (EF$c$) from the fair division literature.
\citet{igarashi2022fair} investigate the interaction of envy-freeness and stability in two-sided matching markets.
These fairness notions seem generally incomparable to ours since they require looking at the potential outcome where one resource was removed (EF1), whereas we guarantee top $k$ outcomes with respect to some baseline.

Among the more traditional stable matching literature which considers uncertainty in inputs, \citet{aziz2016stable} is most similar to our work. They consider producing matchings which are \emph{stable} when the preferences of one side are uncertain in the sense that they are given by a distribution $\D$ over linear orderings of the other side.
We instead only have access to a distribution over merit $\Gamma$, which then induces $\D$, the distribution over linear orderings. 
However, given $\Gamma$, it is not clear that we can exactly construct $\D$ in polynomial time.
Furthermore, the central question in \citet{aziz2016stable} is to find a deterministic matching which maximizes the probability of being stable for the uncertain preferences.
In contrast, we are focused on ensuring good top-$k$ outcomes with respect to some \emph{baseline distribution} over matchings, and are satisfied with randomized outputs.

\paragraph{Statistical Discrimination.} There is also recent work on \emph{statistical discrimination}, which refers to ``discrimination that may occur due to imperfect information a decision maker may have about an individual's qualities'' \citep{castera2022statistical}.
This line of work is relevant to our setting since it implicitly describes situations in which individuals may have high variance in the perceived merit (with respect to the ground truth $\Gamma$).
\citet{emelianov2020fair} initiate the study of candidate selection in the setting where the algorithm only has access to noisy estimates of merit (and the true merit is sampled from normal distributions).
To make a selection, the mechanism is presented with these unbiased merit samples plus some normally distributed noise, where the variance of the noise is group-dependent.
For example, minority groups may have noise with higher variance added to their merit estimates.
\citet{castera2022statistical} extend this inquiry to stable matchings, where they show that different levels of \emph{noise} for different groups of individuals can adversely impact the quality of the matching of \emph{all} individuals, not just the group which the noisy estimates originated from.
We view our work as complementing that of \citet{castera2022statistical}:
instead of observing (noisy) realizations of merit through samples, we assume that the underlying distributions over merits are \emph{known} and accessible to our algorithm.
Even in this full-information case (where we can view the hidden \emph{differential variance} of \citet{emelianov2020fair}), it is unclear
how one should design algorithmic techniques handling distributions, or indeed what even constitutes fairness.
Importantly, in contrast to \citet{castera2022statistical}, we do not make any assumptions on the form of the underlying distributions (e.g., normal distributions).

\paragraph{Uncertainty and Fairness.} More generally, some works do grapple with uncertainty and fairness from a variety of perspectives.
\citet{kearns2017meritocratic, salem2019closing} both focus on uncertainty in fair candidate selection.
The former does so in the context of only being able to evaluate individuals within smaller subsets, and the latter in a secretary problem with uncertainty in applicant quality which is conveyed to the mechanism through partial orders.
\citet{ghosh2021whenfair} investigate the impact of uncertainty in demographic information and its impact on downstream fairness desiderata.

\paragraph{Offline and Online Fair Ranking and Matching.} The extensive literature on fair ranking \cite{singh2018fairness, bower2021individually, celis2020interventions, kletti2022introducing, do2022optimizing, kletti2022pareto, ai2018unbiased} has also influenced our work and ideas; for a comprehensive treatment, see, e.g., \citet{patro2022fairranking}.
Also related is literature on ad markets and marketplaces \cite{celis2019toward, chawla2022individual, ilvento2020multi, basu2020framework, wang2021user}, and the 
growing interest in \textit{online} ranking/matching markets, including fairness considerations \citep{jagadeesan2021learning, jagadeesan2022supply, min2022learn, do2022contextual, patro2020incremental, esmaeili2022rawlsian}.
We focus on the offline setting in this work.

\section{Technical Results}
\label{sec:results}
First, we note that achieving \emph{a} fair solution is straightforward by generalizing the Thompson Sampling approach of \citet{singh2021fairness}.
\begin{proposition}[Thompson Sampling Matching]
\label{prop:thomson-sampling}
  Let $\algthom$ be the following randomized algorithm.
  First, sample a single merit profile $\mv \sim \Gamma$; define rankings $r'_y$ over individuals by the resources according to \cref{prop:fairness_stability_equiv}, i.e., by decreasing merit.
  Then run the Gale-Shapley algorithm using rankings $(r_x)$ and $(r'_y)$, with individuals proposing, and output the resulting matching.
  $\algthom$ is 1-fair.
\end{proposition}
The proof follows from \eqref{eqn:fairness-requirement-after-axiom} since $\algthom$ explicitly produces the same distribution over matchings as defines fairness.

Building on \cref{prop:thomson-sampling}, observe that as in \citet{singh2021fairness}, the principal can achieve $\phi$-fairness with the algorithm $\algmix$
which randomizes between his utility-maximizing matching\footnote{$\M^*$ can be computed in polynomial time as a maximum weighted bipartite matching with respect to the $\mu_{x,y}$.} $\M^*$ (with probability $1-\phi$) and a matching $\M$ obtained from $\algthom$ (with probability $\phi$).
$\algmix$ is $\phi$-fair because $\algthom$ is 1-fair.
Furthermore, $\algmix$ guarantees at least a $1-\phi$ approximation to the optimum utility $U(\M^\star)$ for the principal, since the optimal matching $\M^*$ is chosen with probability at least $1-\phi$.
However, $\algmix$ may not guarantee the optimal utility subject to the approximate fairness constraint.
Indeed, this was already shown by \citet{singh2021fairness} for the special case of rankings.

Our main technical contribution is therefore to derive a \emph{utility-maximizing} algorithm for the principal, in particular in the (realistic) case that the desired fairness approximation $\fm_{x,k}$ must be obtained by sampling.
In \cref{sec:optimal-allocations} we show how to efficiently find the utility-maximizing $\phi$-fair solution with an LP, under the assumption that the parameters $\fm_{x,k}$ are known to the principal.
Then, in \cref{sec:approximate-ranks}, we demonstrate that even when the $\fm_{x,k}$ must be approximated (e.g., by sampling), the principal can obtain near-optimal fairness and utility by suitably modifying the LP; the proof requires a careful perturbation analysis of the LP.

\subsection{Optimal $\phi$-Fair Allocations}
\label{sec:optimal-allocations}

First, we observe a fundamental difference between matching and the ranking case of \citeauthor{singh2021fairness}: in the matching setting, even for $\phi=1$, the fully fair solution is not necessarily unique. (See \cref{sec:multiple-fair} for an example.)
This is in contrast to ranking (when all individuals have the same preference order), where there is a unique solution. 

This is good news for the principal, since, in contrast to ranking, the principal can optimize utility by adjusting allocations even for the most stringent fairness requirement of $\phi=1$.
To characterize the optimal $\phi$-fair solution for the principal, we can write the following linear program $\lpfair$.
In it, the variable $\mpm_{x,y}$ is the probability that individual $x$ is assigned resource $y$. 
The first constraint enforces $\phi$-fairness of the policy induced by $\MPM=(\mpm_{x,y})_{x,y}$; the remaining constraints ensure that $\MPM$ is doubly stochastic.

\begin{LP}[eq:lpfair]{\normalfont{max}}{\sum_{x \in \X} \sum_{y \in \Y} \mu_{x,y} \cdot \mpm_{x,y}}
\sum_{i=1}^k p_{x, r^{-1}_x (i)} \geq \phi \cdot \fm_{x,k} 
& \forall x,k
\\
\sum_{x \in \X} \mpm_{x,y} = 1 
& \forall y 
\\
\sum_{y \in \Y} \mpm_{x,y} = 1 
& \forall x 
\\
\mpm_{x,y} \geq 0
& \forall x, y
\end{LP}

We show a generic and straightforward way to obtain feasible solutions for $\lpfair$ \eqref{eq:lpfair}, thereby also implying that $\lpfair$ is feasible.

\begin{restatable}{proposition}{mfairfeasible}
\label{prop:mfairfeasible}
Assume that for each $x$, the entries $\fm_{x,k}$ form a CDF, in the sense that
$0 = \fm_{x,0} \leq \fm_{x,1} \leq \cdots \leq \fm_{x,n} = 1$, and that for all $y$,
we have $\sum_x (\fm_{x,r_x(y)} - \fm_{x,r_x(y)-1}) = 1$.
For all $x,y$, define $\tfm_{x,y} = \fm_{x, r_x(y)} - \fm_{x, r_x(y) - 1}$.
Then, $\TFM = (\tfm_{x,y})_{x,y}$ is a feasible solution to $\lpfair$ for all $\phi \in [0,1]$.
\end{restatable}
\begin{proof}
  For each $x$, the $\tfm_{x,y}$ form a PDF over the resources $y$; this is directly seen by considering the resources in the order in which $x$ ranks them.
  This immediately implies non-negativity and stochasticity for each $x$ (i.e., each row sums to 1).
  Stochasticity for each $y$ follows from the assumption about the sum over $x$.
  The fairness constraint, even for $\phi=1$, follows by telescoping the sum in the first constraint of $\lpfair$, which equals $\fm_{x,k}$.
\end{proof}

Using a general LP solver for this problem may not always be efficient, especially as the number of individuals $n$ increases.
In \cref{sec:remark_flow}, we show that the optimization can be cast as a weighted (fractional) matching problem.

\subsection{Using Approximate Rank Estimates}
\label{sec:approximate-ranks}

We first show that directly substituting sampled \emph{estimates} of $\fm_{x,k}$ in place of the unknown $\fm_{x,k}$ in the LP may lead to an unfair solution.
Driven by the insight of the failure mode, we then propose an approach for sampling, modifying and solving an LP to avoid it (\cref{prop:sampling_procedure}). We then provide an analysis of the multiplicative\footnote{Note that standard LP sensitivity analysis techniques (e.g., \citet{Cook1986SensitivityTI,nemhauser:wolsey:combinatorial-optimization}) in contrast only yield an additive approximation on the obtained utility, and the main challenge in the analysis is to obtain a multiplicative one.} fairness and utility guarantees of our proposed approach (\cref{thm:main_result}).
Finally, we show that the analysis of our method is tight (\cref{prop:tightness}).

In order to solve $\lpfair$, one needs access to the values $\fm_{x,k}$, which form
the fairness constraints.
Recall that $\fm_{x,k}$ is the probability that individual $x$ is given a resource among her top $k$ preferences when all merits are drawn jointly from the distribution $\Gamma$ and an individual-optimal stable matching is computed with respect to the induced rankings.
Even when $\Gamma$ is given in closed form, with independent and simple distributions for the merits of different individuals and resources, it is not clear how the $\fm_{x,k}$ can be computed exactly; indeed, we believe that it may be \#P-hard to do so. 

The natural alternative 
is to estimate the $\fm_{x,k}$ by sampling from $\Gamma$. 
However, doing so means that the estimates of $\fm_{x,k}$ will not be equal to the true values, which in turn raises the question of how unfair or suboptimal the solution of the LP with perturbed right-hand side could be.
Stated differently, it might take an exorbitant number of samples (and with it time) to obtain sufficiently accurate estimates to ensure approximate fairness and optimality.
To carry out such a sensitivity analysis, we begin with a proposition relating the number of samples with the maximum additive error in approximating the $\fm_{x,k}$.

\begin{restatable}{proposition}{samplingProcedure}
\label{prop:sampling_procedure}
Let $\epsilon > 0$ and $\kappa > 0$ be given, and define $m(\epsilon) = \frac{(\kappa + 1) \cdot \log(2n)}{2\epsilon^2}$.
Let $\usf_{x,k}$ be obtained by sampling merits $\mv$ from $\Gamma$ independently $m(\epsilon)$ times, computing the fair matchings, and normalizing the counts of the outcomes.
With probability at least $1 - n^{-\kappa}$, the resulting estimated fairness requirements $\usf_{x,k}$ 
satisfy $|\usf_{x,k} - \fm_{x,k}| \leq \epsilon$, simultaneously for all $x \in \X$ and $k \in [n]$.
\end{restatable}
\begin{proof}
  The proof is virtually identical to the proof of Proposition~4.3 of \citet{singh2021fairness}.
  Fix some individual $x$.
  Notice that $q_k = \fm_{x,k}$ forms a CDF of the ranks of resources assigned to individual $x$.
  Let the indicator random variable $Z_{k,j}$ be defined by the following process.
  Sample merits $\mv \sim \Gamma$; this sampling is independent for different $j$.
  Define the rankings $r'_y$ for resources as in \cref{prop:fairness_stability_equiv}, and consider the stable matching according to the $r_x, r'_y$ that is optimal for the individuals.
  Let $\mv$ be the resulting stable matching.  
  Let $Z_{k,j} = 1$ iff individual $x$ is assigned a resource she ranks \Kth{k} or better in $\mv$.
  Then $\Pr[Z_{j,k} = 1] = q_k$, and further $Z_k = \frac{1}{m} \cdot \sum_{j=1}^m Z_{k,j}$ is the average of $m$ independent $\text{Bin}(q_k)$ binary random variables.
  By the DKW Inequality for the uniform convergence of the empirical CDF to the true CDF \citep{dvoretzky1956asymptotic}, with probability at least $1 - n^{-\kappa}$, all $\fm_{x,k}$ are estimated with additive error at most $\pm \epsilon$ with $m = \frac{(\kappa + 1)\log(2n)}{2\epsilon^2}$ samples.
\end{proof}

The ``obvious'' way of using the estimates $\usf_{x,k}$ would be to substitute them into $\lpfair$ in place of the unknown $\fm_{x,k}$. 
However, doing so may fail to satisfy any kind of fairness guarantee.
To see this, consider an individual $x$ who has the highest merit for her top choice resource with some small but non-zero probability $\delta < 1/m$ (where $m$ is the number of samples).
There is a non-trivial probability that $\usf_{x,1}$ is estimated to be 0,
so the LP may output a solution $\MPM$ with $\mpm_{x,1} = 0$, completely violating the fairness requirement with respect to individual $x$ and her top resource.
Not only could fairness be completely violated (recall that our fairness requirement is multiplicative), but the principal may even suffer a big loss in utility.
This can occur when $\mu_{x,r^{-1}_{x}(1)}$ is very large, i.e., the principal would derive very high utility from assigning $x$ to her top choice of resource.
The incorrect samples in combination with a stringent fairness requirement (such as $\phi=1$) may prevent the LP from doing so.
Notice that such a large $\mu_{x,r^{-1}_{x}(1)}$ can occur even when $\mu_{x,y} = \Expect{v_{x,y}}$ is the expected merit, namely when $v_{x,y}$ follows a distribution that takes on an extremely large value with very small probability.

Fundamentally, the multiplicative fairness requirement means that the algorithm must guard against rare events of an individual deserving a highly ranked resource.
The approach to ensure this is to allocate to each individual at least a small probability to get high choices, just in case the samples missed the merit values justifying this decision.
More precisely, we define $\nsf_{x,k} := \frac{1}{n\epsilon + 1}(\usf_{x,k} + k\epsilon)$.
Then, the sampling-based algorithm can be summarized as follows:
\begin{enumerate}
    \item Draw $m(\epsilon/2)$ samples from $\Gamma$ to estimate the $\fm_{x,k}$ as $\usf_{x,k}$;
    \item Compute $\nsf_{x,k}$ from the $\usf_{x,k}$ according to $\nsf_{x,k} = \frac{1}{n\epsilon + 1}(\usf_{x,k} + k\epsilon)$;
    \item Solve the LP \eqref{eq:lpfair} with the $\nsf_{x,k}$ in place of the $\fm_{x,k}$, resulting in a marginal probability matrix $\MPM$; and
    \item Compute a Birkhoff von Neumann decomposition of $\MPM$ to obtain a distribution over matchings.
\end{enumerate}

Denote the resulting modified LP, with right-hand side $\nsf_{x,k}$ by $\lpapprox$.
Our main technical result shows that $\lpapprox$ is feasible and gives a good approximation to the fairness and utility guarantees of $\lpfair$.

\begin{restatable}{theorem}{mainResult}
\label{thm:main_result}
  Let $\phi \in [0,1]$ and $\epsilon>0$ be given.
  Assume that the estimates $\usf_{x,k}$ satisfy $|\usf_{x,k} - \fm_{x,k}| \leq \epsilon/2$ for all $x,k$, and define $\nsf_{x,k}$ according to $\nsf_{x,k} = \frac{1}{n\epsilon + 1}(\usf_{x,k} + k\epsilon)$.
  Then $\lpapprox$ is feasible.
  Furthermore, let $\widehat{P}$ be an optimal solution for $\lpapprox$, and $P^*$ an optimal solution for $\lpfair$.
  Then, $\widehat{P}$ is $\left(\frac{\phi \cdot (1 + \epsilon/2)}{n\epsilon+1}\right)$-fair (with respect to the true $\fm_{x,k}$) and has utility $U(\widehat{P}) \geq \frac{1}{\phi n \epsilon + 1} \cdot U(P^*)$.
\end{restatable}
\begin{proof}
    As a first step, we verify that $\lpapprox$ is feasible, and show how to generically obtain a solution.
    With the definition $\nsf_{x,0} := 0$, notice that for any fixed $x$, the $\nsf_{x,k}$ form a CDF.
    This is because the $\usf_{x,k}$ are a CDF (and hence monotone).
    Furthermore, for any resource $y$,
    \begin{align*}
      \sum_x & (\nsf_{x,r_x(y)} - \nsf_{x,r_x(y) -1})
      \\ & = \frac{1}{n\epsilon+1} \cdot \sum_x (\usf_{x,r_x(y) } + \epsilon - \usf_{x,r_x(y) -1})
      \\ & = \frac{1}{n\epsilon+1} \cdot \left( n \epsilon + \sum_x (\usf_{x,r_x(y)} - \usf_{x,r_x(y) -1}) \right)
      \\ & = \frac{1}{n\epsilon+1} \cdot ( n \epsilon + 1 )
      \; = \; 1.
    \end{align*}
    This is because the $\usf_{x,k}$, being obtained by a distribution over matchings, form a doubly stochastic matrix themselves.
    As a result, we can apply \cref{prop:mfairfeasible} with the $\nsf_{x,k}$ in place of $\fm_{x,k}$, and obtain that the solution 
    $\tam_{x,y} := \frac{1}{n\epsilon+1} \cdot (\usf_{x,r_x(y)} - \usf_{x,r_x(y)-1} + \epsilon)$ 
    is a feasible solution to $\lpapprox$.

  It remains to prove fairness and utility guarantees; 
  We begin with fairness.
  Let $P = (p_{x,y})_{x,y}$ be a feasible solution to $\lpapprox$.
  Consider some individual $x$ and $k \in \SET{1, \ldots, n}$.
  Because $P$ is a $\phi$-fair solution of $\lpapprox$, it satisfies
  $\sum_{i=1}^k p_{x,r_x^{-1}(i)} \geq \phi \cdot \nsf_{x,k} = \frac{\phi}{n\epsilon + 1}(\usf_{x,k} + k\epsilon)$.
  By assumption, $\usf_{x,k} \geq \fm_{x,k} - \epsilon/2$, which together with $k \geq 1$ and $\fm_{x,k} \leq 1$ implies that
  $\frac{\phi}{n\epsilon + 1} (\usf_{x,k} + k \epsilon) \geq \frac{\phi}{n\epsilon + 1} (\ell_{x,k} + (\epsilon/2) \cdot \ell_{x,k} ) \geq \frac{\phi \cdot (1 + \epsilon/2)}{n\epsilon + 1} \fm_{x,k}$,
  so we have established $\left(\frac{\phi \cdot (1 + \epsilon/2)}{n\epsilon + 1}\right)$-fairness.
  The remainder of the proof --- and most of the technical work --- will be concerned with the loss in the principal's utility.

   Let $P^* = (p^*_{x,y})_{x,y}$ be an optimal solution to $\lpfair$.
   Define the matrices $\TFM = (\tfm_{x,y})_{x,y}, \TAM = (\tam_{x,y})_{x,y}$ as before as
   \begin{align*}
     \tfm_{x,y} & := \fm_{x,r_x(y)} - \fm_{x,r_x(y)-1}
     & 
     \tam_{x,y} & := \frac{1}{n\epsilon+1} \cdot (\usf_{x,r_x(y)} - \usf_{x,r_x(y)-1} + \epsilon).
   \end{align*}
   By \cref{prop:mfairfeasible}, $\TFM$ constitutes a feasible solution for $\lpfair$, and $\TAM$ a feasible solution for $\lpapprox$.
  We first show that the entries of $\TAM$ cannot be much smaller than those of $\TFM$:
  \begin{align}
    (1 + n \epsilon) \cdot \tam_{x,y} - \tfm_{x,y} & \geq 0 \qquad \text{ for all } x, y.
            \label{eqn:tfm-tam-close}
  \end{align}
  To do so, we recall that by assumption of the theorem, $|\usf_{x,k} - \fm_{x,k}| \leq \epsilon/2$ for all $x, k$, so by definition of $\tfm_{x,y}$ and $\tam_{x,y}$, we have
  \begin{align*}
    (1 + n \epsilon) \tam_{x,r_x(y)} - \tfm_{x,r_x(y)}
    & = (\usf_{x,k} - \usf_{x,k-1} + \epsilon) - (\fm_{x,k} - \fm_{x,k-1})
    \; = \; \epsilon + \underbrace{(\usf_{x,k} - \fm_{x,k})}_{\geq -\epsilon/2} - \underbrace{(\usf_{x,k-1} - \fm_{x,k-1})}_{\leq \epsilon/2}
      \; \geq \; 0.
  \end{align*}

    Now, we define the matrix
    \begin{align*}
      W & = \frac{\phi n \epsilon + \phi}{\phi n \epsilon + 1} \cdot \TAM + \frac{1}{\phi n \epsilon + 1} \cdot P^* - \frac{\phi}{\phi n \epsilon + 1} \cdot \TFM.
    \end{align*}
    Being feasible solutions to the respective LPs, $\TAM, P^*, \TFM$ are all doubly stochastic, and in particular have row and column sums 1. Therefore, $W$ is a linear combination of matrices with row and column sums 1, and the respective coefficients of $\TAM, P^*, \TFM$ are $\frac{\phi n \epsilon + \phi}{\phi n \epsilon + 1}, \frac{1}{\phi n \epsilon + 1}, \frac{-\phi}{\phi n \epsilon+1}$ and sum up to 1.
    Thus, by linearity, $W$ also has row and column sums 1.

    Next, we show that the entries of $W$ are all non-negative. 
    Using the definition of $W$ and \eqref{eqn:tfm-tam-close}, we can bound
  \begin{align*}
    w_{x,y} & = \frac{\phi n \epsilon + \phi}{\phi n \epsilon + 1} \cdot \tam_{x,y}
              + \underbrace{\frac{1}{\phi n \epsilon + 1} \cdot p^*_{x,y}}_{\geq 0}
              - \frac{\phi}{\phi n \epsilon + 1} \cdot \tfm_{x,y}
       \; \geq \; \frac{\phi}{\phi n \epsilon+1} \cdot ( (1 + n\epsilon) \cdot \tam_{x,y} - \tfm_{x,y} )
       \; \geq 0.
  \end{align*}

  Thus, we have shown that $W$ is doubly stochastic.   
  Next, we will show that $W$ also satisfies the fairness constraints of $\lpapprox$, and is thus feasible for $\lpapprox$.
  To do so, fix an $x$ and a $k$. Using the definition of $W$, we can write

    \begin{align*}
      \sum_{i=1}^k w_{x, r_x^{-1}(i)}
      & = \underbrace{\frac{\phi n \epsilon + \phi}{\phi n \epsilon + 1}}_{\geq \phi} \cdot \sum_{i=1}^k \tam_{x,r_x^{-1}(i)} + \underbrace{\frac{1}{\phi n \epsilon + 1} \cdot \sum_{i=1}^k (p^*_{x,r_x^{-1}(i)} - \phi \tfm_{x,r_x^{-1}(i)})}_{(*)}.
    \end{align*}

    We now show that the expression $(*)$ is non-negative.
    Recall that the fact that $P^*$ satisfies the $\phi$-fairness requirement, along with the definition of $\TFM$ (and the resulting telescoping series) implies that

    \begin{align*}
      \sum_{i=1}^k p^*_{x,r_x^{-1}(i)}
      & \geq \phi \cdot \fm_{x,k}
      \; = \; \phi \cdot \sum_{i=1}^k \tfm_{x,r_x^{-1}(i)}.
    \end{align*}

    Using this non-negativity and the definition of $\TAM$, we obtain that

    \begin{align*}
      \sum_{i=1}^k w_{x, r_x^{-1}(i)}
      & \geq \phi \cdot \sum_{i=1}^k \tam_{x,r_x^{-1}(i)}
      \; = \; \phi \cdot \nsf_{x,k},
    \end{align*}
    thus implying that $W$ is a feasible solution to the approximate LP.

    Given that $W$ is a feasible solution for $\lpapprox$, its utility gives a lower bound on the utility of optimal solutions to $\lpapprox$.
    We therefore complete the proof by lower-bounding the utility achieved from $W$.
    
    \begin{align*}
       U(W) & = \sum_{x \in \X} \sum_{y \in \Y} \mu_{x,y} \cdot w_{x,y}
        \; = \; \frac{1}{\phi n \epsilon + 1} \cdot 
             \sum_{x \in \X} \sum_{y \in \Y} \mu_{x,y} \left( p^*_{x,y} + \phi \cdot \underbrace{\left( (n \epsilon + 1) \cdot \tam_{x,y} - \tfm_{x,y}\right)}_{\geq 0 \text{ by } \eqref{eqn:tfm-tam-close}} \right)
        \; \geq \; \frac{1}{\phi n \epsilon + 1} \cdot U(P^*).
    \end{align*}
\end{proof}

Theorem \ref{thm:main_result} gives the desired multiplicative approximation on the utility obtained by the modified LP. 
Combining \cref{thm:main_result} with \cref{prop:sampling_procedure}, we obtain that by sampling from $\Gamma$
$O(\log (n)/\epsilon^2)$ times, then solving the resulting linear program $\lpapprox$, with high probability, the output will be at least $\frac{\phi(1 + \epsilon/2)}{n \epsilon + 1}$ fair and approximate the principal's optimal utility to within a factor of $\frac{1}{\phi n \epsilon + 1}$.

While our result can be considered an extension/generalization of Proposition~4.4 from \citet{singh2021fairness}, several points are worth mentioning. 
First, our setting is significantly more general, and the fact that individuals do not have identical preferences requires a much more careful proof approach.
Second, our \cref{thm:main_result}  improves the approximation guarantee for the principal's utility from $\frac{1}{n\epsilon+1}$ to $\frac{1}{\phi n \epsilon + 1}$. While this may appear to be a small improvement, it is quite meaningful: in particular, for $\phi=0$,
we recover the fact that an unconstrained principal does not suffer any utility loss from misestimating the fairness requirements (which will not be enforced anyway).
Finally, the proof of Proposition~4.4 in \citet{singh2021fairness} in its current form contains a serious mistake; indeed, it appears that a proof along the lines pursued in \citet{singh2021fairness} cannot succeed, so our work is the first correct proof of the claimed result.

The analysis in \cref{thm:main_result} is tight: for the given algorithm, there are instances for which perturbed inputs lead to the given loss in fairness and utility.
\begin{restatable}{proposition}{tightnessResult}
    \label{prop:tightness}
    There exists a fair ranking instance for $\phi=1$ and a small perturbation for which the algorithm of \cref{thm:main_result} only achieves a $\frac{1}{n \epsilon + 1}$ approximation for utility, and another instance for which it only achieves fairness $\frac{1+\epsilon/2}{n\epsilon+1}$.
\end{restatable}
\begin{proof}
We explicitly construct the true merit distributions and small adversarial perturbations, for $n$ individuals.
Our examples in fact fit within the \emph{ranking} (not just the \emph{matching}) framework, in that for any given individual $i$, the merit for all resources is the same. We therefore simply talk about the merit of each individual, without referencing a resource.

In the example showing tightness of the guarantee on the principal's utility, 
individual 1 has merit $n$ with probability $1-\epsilon$ and merit $n-1$ with probability $\epsilon$.
Individual 2 has merit $n-1$ with probability $1-\epsilon$ and merit $n$ with probability $\epsilon$.
Each other individual $i$ has merit $n-i$ deterministically, i.e., with probability 1.
The PDF $p_\ell$ and CDF $\ell$ are therefore

\begin{align*}
    &p_\ell = \begin{bmatrix}
        1-\epsilon & \epsilon & 0 \\
        \epsilon & 1-\epsilon & 0 \\
        0 & 0 & I_{n-2} \\
    \end{bmatrix}
    &\ell = \begin{bmatrix}
        1 - \epsilon & 1 & 1 & \dots & 1\\
        \epsilon & 1 & 1 & \dots & 1\\
        0 & 0 & 1 & \dots & 1\\
        \vdots & \vdots & \ddots & \ddots & \vdots \\
        0 & 0 & \dots & 0 & 1
    \end{bmatrix}.
\end{align*}
Here, $I_{n-2}$ is the $n-2 \times n-2$ identity matrix.
According to this distribution, the ranking of individuals by merit is the same as their ranking by index, except for some randomization between individuals 1 and 2.

Next, we define an adversarially perturbed version of the PDF and CDF as follows: 
\begin{align*}
    &p_{\usf} = \begin{bmatrix}
        1-\epsilon/2 & 0 & 0 & \dots & \epsilon/2\\
        \epsilon/2 & 1 - \epsilon/2 & 0 & \dots & 0\\
        0 & \epsilon/2 & 1-\epsilon/2 &\dots & 0\\
        \vdots & \vdots & \vdots & \ddots & \vdots\\
        0 & \dots & \dots & \epsilon / 2 & 1-\epsilon/2
    \end{bmatrix}
    &\usf = \begin{bmatrix}
        1 - \epsilon/2 & 1 - \epsilon/2 & 1 - \epsilon/2 & \dots & 1\\
        \epsilon/2 & 1 & 1 & \dots & 1\\
        0 & \epsilon/2 & 1 & \dots & 1\\
        \vdots & 0 & \ddots & \ddots & \vdots \\
        0 & 0 & \dots & \epsilon/2 & 1
    \end{bmatrix}.
\end{align*}

Under this perturbed distribution, individual 1 has merit $n$ with probability $1-\epsilon/2$, and otherwise has merit 0,
individual 2 has merit $n$ with probability $\epsilon/2$, and otherwise has merit $n-1$,
individual 3 has merit $n-1$ with probability $\epsilon/2$, and otherwise has merit $n-2$, etc.
In particular, note that the first row of $\usf$ is $1-\epsilon/2$ everywhere except for the last column, where it is 1. 
By inspecting the CDFs of both distributions, we see that the difference in any entry is at most $\epsilon/2$, so these distributions indeed satisfy the assumptions of \cref{thm:main_result}.

Finally, we define the principal's utility function: it is $\mu_{x,k} = 0$ everywhere for all $x, k$ except for $\mu_{1, 2} = 1$.
That is, the principal obtains utility only if individual 1 is ranked in exactly the second spot $k=2$.
(If we view the spots as jobs which all individuals rank in the same order, this would correspond to individual 1 being a particularly good fit with job 2.)

Using the definition $\nsf_{x,k} = \frac{1}{n\epsilon + 1}(\usf_{x,k} + k\epsilon)$, we can now calculate.
\begin{align*}
    \nsf_{1,1} &= \frac{1}{n\epsilon + 1}(\usf_{1,1} + \epsilon) 
    = \frac{1 + \epsilon/2}{n\epsilon+1}
    & 
    \nsf_{1,2} &= \frac{1}{n\epsilon + 1}(\usf_{1,2} + 2\epsilon) 
    = \frac{1 + \tfrac{3}{2}\epsilon}{n\epsilon + 1}.
\end{align*}
For $\phi=1$, in ranking, the solution is unique (see \cref{sec:multiple-fair} and proof of Lemma~4.2 by \cite{singh2021fairness}), and given by the PDFs induced by the CDFs.
In particular, we obtain that the probability of allocating individual 1 to slot 2 (the only allocation resulting in any utility) is
\begin{align*}
    \widehat{p}_{1,2} = \nsf_{1,2} - \nsf_{1,1} = \frac{1 + \tfrac{3}{2}\epsilon - 1 - \tfrac{1}{2}\epsilon}{n\epsilon+1}
    = \frac{\epsilon}{n\epsilon + 1}.
\end{align*}
However, the allocation given by the true 1-far PDF $Q = p_{\ell}$ has $q_{1,2} = \epsilon$.
Therefore, the utility under $\ell$ is $\epsilon$, whereas the utility under $\nsf$ is $\frac{\epsilon}{n\epsilon + 1}$.
This proves that the $\frac{1}{n\epsilon + 1}$ approximation ratio for the utility is tight.

Next, we modify the example slightly to obtain a tight example for the approximation of fairness. Here, the ground truth is even simpler: the PDF is the identity matrix $I_n$, i.e., individual $i$ deterministically has the $i$-th highest merit.
We use the same perturbed input $\usf$ and $p_{\usf}$ as before. Notice that it also perturbs each value by at most $\epsilon/2$ compared to this new ground truth $I_n$.
As we showed above, $\nsf_{1,1} = \frac{1 + \epsilon/2}{n\epsilon+1}$, whereas according to the ground truth, individual 1 is always the best (and thus entitled to the top spot). 
Due to the uniqueness of the solution in the context of ranking for $\phi=1$, we have that $\widehat{p}_{1,1} = \nsf_{1,1} = \frac{1 + \epsilon/2}{n\epsilon+1} \cdot \ell_{1,1}$.
Thus, on this example, the distribution achieves fairness no better than $\frac{1 + \epsilon/2}{n\epsilon+1}$, completing the proof.
\end{proof}

\section{Experiment}
\label{sec:experiments}
To evaluate how closely the practical performance aligns with our theoretical results, and how our method compares with prior work, we ran a simple experiment.
The experiment uses a public dataset from the Czech dating site \textit{Libimseti} \citep{brozovsky07recommender} which has been used in experiments in prior work on matching (e.g., \citet{su2022optimizing}). 
For each user, the dataset contains a binary attribute of their gender, as well as ratings by other users of the opposite gender, on a scale of [1-10]. 
To avoid any normative connotations of the side of the market that should be treated fairly, we will refer to the two sides as ``orange'' and ``green'' instead of gendered.

For computational reasons, our experiment uses a subset of the data consisting of the 100 highest contributing orange and green users (those who have rated the most users of the opposite color).
Based on these 200 users, we construct two $100 \times 100$ matrices. 
The first, $R_G$, has at entry $R_G(x,y)$ the rating of orange individual $y$ by green individual $x$.
Similarly, $R_O(x,y)$ gives the rating of green individual $y$ by orange individual $x$. 
We impute missing entries (i.e., when user $x$ had not rated user $y$ in the true data set) in either matrix using a standard matrix completion technique \citep{salakhutdinov2008bayesian}. 

In our problem setup, one side of the market has certain preferences (i.e., rankings) while the other has uncertain estimates of the merits of the first side. To model preferences, we take as ground truth the matrix completion of $R_G$, and use it to get a deterministic ranking function over the entire set of orange individuals for each green individual. 
To model uncertainty in the merits, we assume that the merits follow a normal distribution centered on the estimated rating: $r(x,y) \sim \mathcal{N}(R_O(x,y), \sigma = 3)$, where 
$r(x,y)$ is the rating (merit) of the $y$th green user by the $x$th orange user.

We treat $\mu_{x,y} = \Expect{r(y,x)} = R_O(y,x)$ as the mean rating that green user $x$ receives from orange user $y$, and our optimization objective $\sum_{x=1}^{100} \sum_{y=1}^{100} \mu_{x,y} p_{x,y}$ therefore maximizes the expected welfare of the orange people. 
(Note that the reversed index in $\mu_{x,y}$ is because in the LP \eqref{eq:lpfair}, each variable $p_{x,y}$ represents the probability that green person $x$ is matched with orange person $y$.) 
To construct the approximate RHS fairness constraints $\nsf$ in the linear program \cref{eq:lpfair}, we ran the Gale-Shapley algorithm 10,000 times, each time with different merits sampled from the aforementioned rating distributions.
We set $\epsilon = 0.01$, so 10,000 samples is approximately sufficient to obtain error at most $\epsilon/2$ by \cref{prop:sampling_procedure}.

\begin{figure}
    \centering
    \includegraphics[scale=0.52]{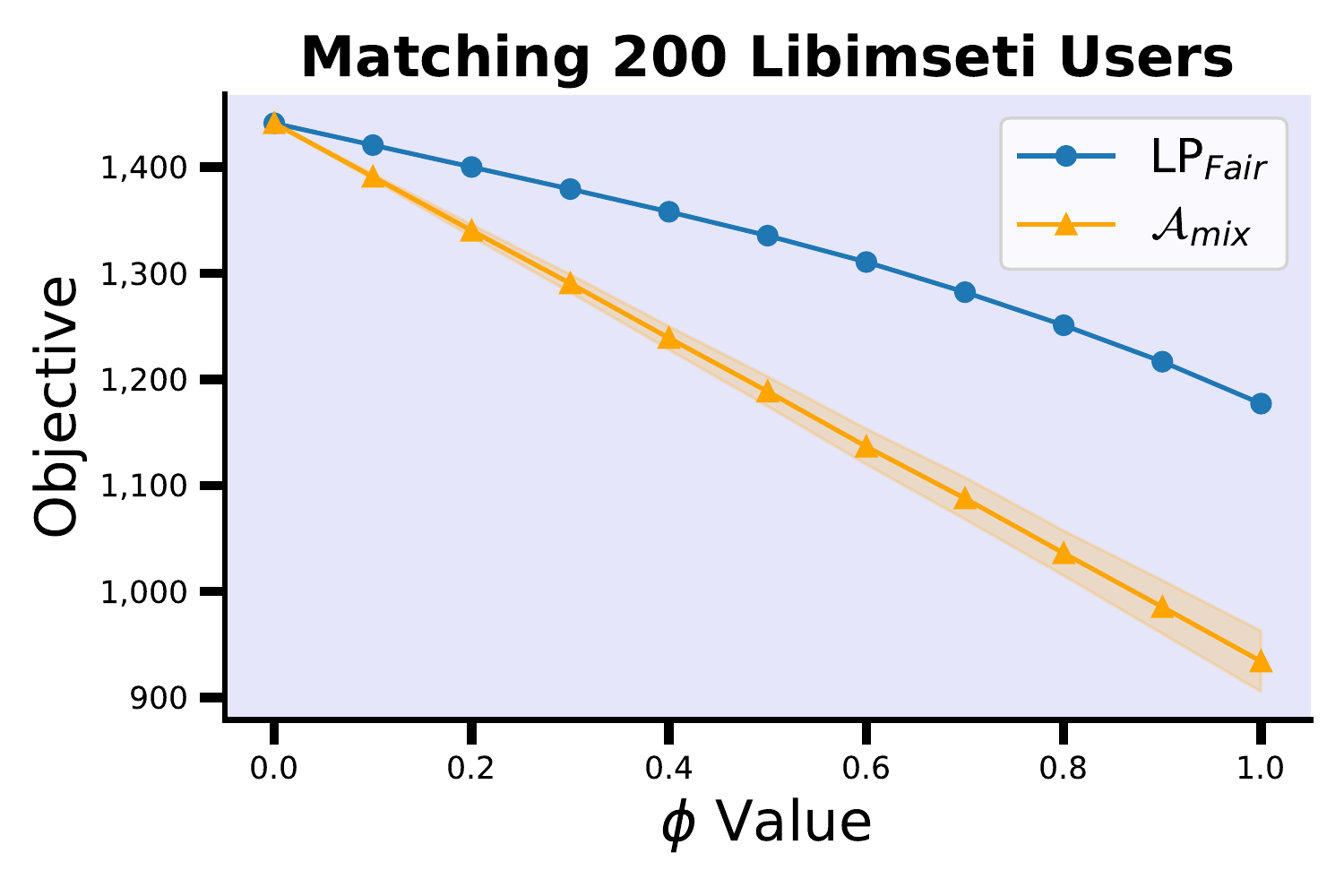}
    \caption{$\phi$ vs.~utility value for experiments on the Libimseti dating dataset \cite{brozovsky07recommender}. Our method (blue) obtains higher utility than the baseline $\algmix$ Thompson Sampling (orange) everywhere except $\phi=0$ (an unconstrained principal).
    We obtain a performance boost even for 1-fairness, which was not the case in previous work \citep{singh2021fairness}.}
    \label{fig:experiment_results}
\end{figure}

\cref{fig:experiment_results} depicts the drop in utility across different $\phi$ values for the LP-based solution $\lpapprox$, vs.~the baseline which randomizes, with parameter $\phi$, between Thompson sampling and a utility-maximizing matching.

As in \cite{singh2021fairness} observe that $\algmix$ is a straight line, since, in expectation, it is just a convex combination of two different objective values.
Notice that in our setting, $\lpapprox$ actually achieves higher utility than Thompson sampling at $\phi=1$. 
This is in contrast to the setting of ranking (and the MovieLens experiments from \cite{singh2021fairness}, Figure 2b). It happens because in matching, there may be different 
$1$-fair solutions (see, e.g.,  \cref{sec:multiple-fair}). 
Our experiments thus suggest that in the context of matching, there may be a much larger margin for potential utility optimization while preserving the same level of fairness, in particular for high values of $\phi$-fairness. This is because when the preferences of individuals are not aligned, they may be less in competition for resources.

\section{Extensions}
\label{sec:extensions}
\paragraph{Beyond One-to-one Matchings.} For ease of exposition, we assumed that $|\X| = |\Y| = n$, i.e., the number of resources equals the number of individuals.
This assumption was not crucial at all, as we now discuss.
As is standard in discussing allocation preferences, we assume that every individual $x$ prefers being allocated to any resource $y$ over not being allocated at all.
First, notice that \cref{axiom:fairness_certainty} did not require the number of individuals and resources to be the same --- it simply states a requirement for each pair $(x,y)$ of an individual and resource. 
Correspondingly, the same fairness statistics defined in \cref{sec:preliminaries} still capture the intuitive notion of fairness, i.e., the frequencies/probabilities of individual $x$ being allocated rank $k$ or higher. 
Consequently, \cref{eqn:fairness-requirement-after-axiom} is still a meaningful fairness requirement under uncertainty.

The primary difference is that some individuals or resources may go unallocated, so the allocation matrices will typically not be doubly stochastic.
Also, the equivalence between fair matchings and stable matchings with respect to induced rankings $r'_y$ by resources needs a minor adaptation.
Following the standard approach in the area (see Section~1.4.1 of \cite{gusfield1989stable}), we can extend the definition of stability to instances with unequal sides to allow for unmatched individuals/resources.
A standard reduction involving the introduction of additional ``virtual individuals/resources'' who are ranked lowest (in arbitrary order) by everyone on the other side, and whose own rankings are irrelevant, can be used to find stable matchings in this context.

In summary, we still obtain an equivalence between fair matchings and stable matchings, which can be combined with \cref{axiom:lifting} to obtain all relevant statistics and define a suitable linear program.
The rest of the approach and analysis carry through unchanged.
The easiest way to see this is to again introduce ``virtual individuals/resources'' and reduce the case of $|\X| \neq |\Y|$ to the case of $|\X| = |\Y|$, defining the principal's utility $\mu_{x,y} = 0$ when the individual $x$ or resource $y$ is virtual.
A similar approach can be taken in the setting where resources may be able to accept multiple individuals.

\paragraph{Beyond Linear Objectives.} Another possible generalization would be to consider non-linear objective functions for the principal. (Recall from \cref{sec:principal-utility} that we assumed the principal's utility to be the sum of utilities for each match between an individual and a resource.)
For non-linear objectives, the primary problem is that the principal's utility cannot be fully captured by the marginal probabilities $p_{i,j}$ with which each resource is allocated to each individual. 
Therefore, it is not clear how to formulate a tractable optimization problem in this setting.
As a simple example, consider the case when the principal's objective is to maximize the minimum utility of any agent. When there are two agents and one desirable item, the minimum utility will always be 0 (determined by the agent who does not receive the item). However, a fractional allocation ---  captured by the LP --- will consider allocating ``half'' of the item to each agent, obtaining positive minimum utility.
While one could still find an optimum distribution by writing an LP with one variable for each full allocation, this approach results in an exponentially large linear program, and will thus typically not be practical or useful.

\section{Future Work}
\label{sec:future-work}
Recall that we assumed access to the merit distribution $\Gamma$, which we treat as the underlying ground truth. A ripe avenue for future work is to consider how to actually obtain $\Gamma$, or indeed a good enough approximation $\widehat{\Gamma}$, to run our procedure and still obtain guarantees with respect to the underlying ground truth.
Similarly, it may be non-trivial to obtain the individuals' full rankings, or to determine how much information about the rankings is necessary to arrive at the same outcomes as our approach.

It is also natural to apply our framework to other allocation problems, or possibly beyond.
Here, the fact that our technical results (apart from the perturbation analysis) were extremely straightforward is very promising.
For it suggests that when viewed within the right framework, fairness requirements may not pose a big obstacle, so a generalization to more challenging settings may be within reach.

The broader point raised by our work is that fairness may be very fruitfully modeled and analyzed by quantifying uncertainties in the merit predictions of any ML algorithm or domain expert.
Our work clearly highlights that achieving fairness in allocation relies on learning accurate posterior distributions $\Gamma$ from the given observable variables.
It shifts the conversation from ``when should we consider individuals similar?'' to ``what do observable features tell us about individuals' merits?'', a framing that may be much more clearly understood by domain experts, and is more amenable to standard ML approaches.
Understanding the implications of this requirement and developing domain-specific approaches to meet it is a promising direction for future work.

\subsubsection*{Acknowledgements}
SD was supported by the Department of Defense (DoD) through the National Defense Science \& Engineering Graduate (NDSEG) Fellowship Program.
This work was also funded in part by NSF awards \#1916153, \#1956435, \#1943584, and \#2239265.

\bibliographystyle{abbrvnat}
\bibliography{davids-bibliography/names,davids-bibliography/conferences,davids-bibliography/bibliography,davids-bibliography/publications,references}

\newpage
\appendix

\section{An Example with Multiple 1-Fair Allocation Distributions}
\label{sec:multiple-fair}
Here, we show that already for $n=3$ individuals and resources, there are instances which not only have multiple 1-fair distributions over matchings, but even have distributions over matchings which have different \emph{marginal probabilities} for allocating specific resources to individuals.
This constitutes a stark contrast with the setting of ranking, in which there is a unique 1-fair distribution over the marginals of rankings.

In our instance, the individuals' preference rankings are given by the following --- for example, individual 1 ranks the resources in the order $(1,3,2)$ from most to least desirable: 
\begin{align*}
    \pref_\X & = \begin{bmatrix}
        1 & 3 & 2\\
        1 & 3 & 2\\
        3 & 1 & 2
    \end{bmatrix}. 
\end{align*}

The distribution of merits is given by 
\begin{align*}
    \Gamma = 
    \begin{bmatrix}
    0 & 0 & 1\\
    1 & 1 & 2\\
    2 & 2 & 0
    \end{bmatrix}
    \text{ with probability } 0.9, \qquad  \text{ and }
    \begin{bmatrix}
        2 & 0 & 0\\
        1 & 1 & 1\\
        0 & 2 & 2
    \end{bmatrix}
    \text{ with probability } 0.1;
\end{align*}
again, the rows correspond to individuals, so in the first (more likely) case, individual $1$ is least meritorious for resources 1 and 2, and in the middle for resource 3.
Notice that the distribution is such that there are never ties in merit \emph{for the same resource}.
By inferring the corresponding rankings $r'_y$ by the resources induced from the merits, and computing the individual-optimal stable matchings, we obtain that in the two cases, the individual-optimal stable matchings are
\begin{align*}
    M_1 & = \begin{bmatrix}
        0 & 1 & 0\\
        0 & 0 & 1\\
        1 & 0 & 0
    \end{bmatrix}
    & M_2 & = \begin{bmatrix}
        1 & 0 & 0\\
        0 & 1 & 0\\
        0 & 0 & 1
    \end{bmatrix}.
\end{align*}

By observing the position which each individual obtains in her ranking for each of these two allocations, and taking the convex combination, we see that the fairness requirements are captured by the following matrix:

\begin{align*}
  \FM & = (\fm_{x,k})_{x,k}
        \; = \; \begin{bmatrix}
   0.1  & 0.1 & 1\\
   0    & 0.9 & 1\\
   0.1  & 1   & 1
   \end{bmatrix}.
\end{align*}

Now, consider the alternative randomized allocation which chooses
\begin{align*}
   \Tilde{M}_1 & = \begin{bmatrix}
        0 & 1 & 0\\
        1 & 0 & 0\\
        0 & 0 & 1
      \end{bmatrix}
\end{align*}
with probability $0.9$, and the matrix $M_2$ given above with probability $0.1$.
Under this distribution of allocations, the probabilities for each individual $x$ to obtain her \Kth{k} ranked choice are captured by the following matrix:
\begin{align*}
  \FM' & = (\fm'_{x,k})_{x,k}
        \; = \; \begin{bmatrix}
   0.1  & 0.1 & 1\\
   0.9  & 0.9 & 1\\
   1    & 1   & 1
   \end{bmatrix}.
\end{align*}

Because $\fm'_{x,k} \geq \fm_{x,k}$ for all $x$ and $k$, this distribution is 1-fair; furthermore, not only the distribution over matchings is different, but so are the marginals.
Indeed, we can generalize this example to obtain a whole family of 1-fair distributions with distinct marginals of allocations: with probability 0.9, the individual chooses one of $M_1, \Tilde{M}_1$ (choosing the former with some probability $\lambda \in [0,1]$), and with the remaining probability, the principal chooses $M_2$.

It is also worth noting that the matrix $\Tilde{M}_1$ is not stable under the rankings $r'_y$ for either of the two possible merit profiles. Indeed, under the first merit profile, individual 1 and resource 3 form an instability, and under the second merit profile, individual 1 and resource 1 form an instability.
This serves as a counterexample to a potential conjecture that any 1-fair distribution over allocations must randomize over allocations which are stable for \emph{some} merit profile in the support of $\Gamma$.

\section{Solving $\lpfair$ with Weighted Fractional Matching}
\label{sec:remark_flow}
As we mentioned in \cref{sec:results}, while the linear program $\lpfair$ is a natural way to capture the principal's optimization problem, the solution can in fact be obtained without an explicit LP solver, by noticing that the optimization problem can be cast as a weighted (fractional) matching problem.
This is not all that surprising, given that the underlying problem is to match individuals with resources.
Recall that maximum-weight matching (and more generally, circulation) problems can be solved using combinatorial algorithms \citep{ahuja:magnanti:orlin}, which may provide
a more efficient tailored algorithm for larger $n$.

In order to show a reduction to the weighted matching problem, we interpret the probabilities $\mpm_{x,y}$ as fractional assignments or flows in a circulation problem.
For each resource $y$, there is a node $u'_y$, whose total demand is 1.
To capture the fairness requirements $\fm_{x,k}$, for each individual $x$, we have $n$ nodes $u_{x,1}, \ldots, u_{x,n}$.
For $k < n$, node $u_{x,k}$ has supply $\phi \cdot (\fm_{x,k} - \fm_{x,k-1})$; and $u_{x,n}$ has supply $\phi \cdot (\fm_{x,n} - \fm_{x,n-1}) + (1-\phi)$.
The bipartite graph contains exactly the edges $(u_{x,k}, u'_y)$ for $r_x(y) \leq k$, with weights $\mu_{x,y}$.
Note that by this definition, $u_{x,n}$ has edges to all $u'_y$.
The supply at $u_{x,k}$ is exactly the probability with which $x$ must be assigned a rank $k$ or higher to ensure $\phi$-fairness, but excluding the probability with which $x$ must be assigned a rank of $k-1$ or higher.
It is then straightforward (and standard) to verify that the maximum-weight circulation with the given demands and supplies is an optimum solution to $\lpfair$.

Indeed, we are not the first to observe the connection between this type of LP and bipartite circulations. A similar construction was already presented in the first step of Algorithm~1 of \citet{athanassoglou2011house} in a housing allocation problem with probabilistic inputs.

To see why the reduction is useful, recall that maximum-weight matching (and more generally, circulation) problems can be solved using combinatorial algorithms \citep{ahuja:magnanti:orlin}. 
Therefore, combined with noting that this procedure will also work for $\lpapprox$ using $\TAM$ (defined in the proof of \cref{thm:main_result} in \cref{sec:results}) instead of $\TFM$ implies that if we wanted to find a solution satisfying \cref{thm:main_result}, we could just run a combinatorial algorithm (such as a min-cost max-flow) instead of a generic LP solver.

\end{document}